\documentclass{article}
\usepackage[T1]{fontenc} 
\usepackage[utf8]{inputenc} 
\usepackage{amsmath,cite,url}
\usepackage{graphicx}
\usepackage[lbd]{ismir}

\usepackage{color}

\usepackage{booktabs}
\usepackage[firstpage]{draftwatermark}
\definecolor{lightgray}{rgb}{0.9,0.9,0.9}
\definecolor{darkgray}{rgb}{0.4,0.4,0.4}
\SetWatermarkFontSize{12pt}
\SetWatermarkScale{1.1}
\SetWatermarkAngle{90}
\SetWatermarkHorCenter{202mm}
\SetWatermarkVerCenter{170mm}
\SetWatermarkColor{darkgray}
\SetWatermarkText{Late-Breaking / Demo Session Extended Abstract, ISMIR 2025 Conference}

\title{An Optical Music Recognition Tool for String Quartets}

\multauthor
{Jin-Shiang Hu$^{1,3}$ \hspace{1cm} Bo-Yuan Cheng$^{1,2}$ \hspace{1cm} Li Su$^1$} {
 $^1$ Academia Sinica, Taiwan \hspace{0.7cm}
$^2$ National Taiwan University \hspace{0.7cm}
$^3$ University of British Columbia\\
{\tt\small jhu35@student.ubc.ca, b11901020@ntu.edu.tw, lisu@iis.sinica.edu.tw}
}

\def\authorname{J. Hu, B. Cheng, L. Su}

\begin{document}

\maketitle
\begin{abstract}
This paper presents an optical music recognition (OMR) system tailored for PDF scores of string quartet pieces. The approach begins with a SegNet-based segmentation model to localize musical regions, followed by convolutional neural networks and image processing techniques to classify and extract musical symbols. To improve rhythmic precision, dynamic programming methods enforce time signature constraints and explicitly handle tuplets. Evaluation using a longest common subsequence metric across multiple feature matrices demonstrates robust accuracy in recognizing pitch, onset, duration, and accidentals on a benchmark dataset. Code and data are available at \url{https://github.com/lattellie/25-omr}.
\end{abstract}

\section{Introduction}\label{sec:introduction}
Optical music recognition (OMR) \cite{rios2024sheet,calvo2023optical,van2017optical} is a fundamental technique in processing music notation. 
Several challenges remain in OMR, including the alignment of music across multiple instrumental tracks, consistency in bar lengths according to time signatures, and the handling of tuplets and irregular rhythmic patterns. 
For multi-staff OMR, 
string quartets provide a well-structured and manageable test case. 
Despite being more compact, they encompass the full range of clefs, key signatures, double stops, and complex rhythmic patterns. This makes them a suitable intermediate step toward more complex 
scores, allowing us to address the 
challenges in a controlled setting.

Our system focuses on performing OMR on PDF sheet music of string quartet compositions. 
Our approach begins with 
a SegNet model 
taken directly from the open-source repository at  \url{https://github.com/BreezeWhite/oemer} \cite{oemer}. 
We used this pre-trained model, 
which forms the foundation for locating staff regions and isolating individual musical elements. 
Then, we apply a combination of traditional image processing techniques and convolutional neural networks (CNNs) to further refine the detection and classification of individual music symbols. 
Using OpenCV \cite{itseez2015opencv}, we implemented a pipeline to classify musical symbols, infer rhythmic structure, correct timing to enforce time signatures, handle tuplets, and convert the results into an editable MusicXML format.
Also, to address rhythmic accuracy, we incorporate a dynamic programming-based correction step that enforces timing constraints, ensuring that bar lengths are consistent and tuplets are correctly interpreted and represented. 



We utilize the Annotated Beethoven Corpus (ABC) dataset for evaluation \cite{abc}. The number of correctly recognized notes is determined by computing the longest common subsequence (LCS) between the ABC reference and our system’s output. This approach better captures the sequential structure of music, addressing a key limitation of Symbol Error Rate (SER) \cite{ser2023,rios2024sheet,edirisooriya2021empirical,martinez-sevilla2023}, which does not account for the musical order of events.

\begin{figure*}
  \includegraphics[width=\textwidth]{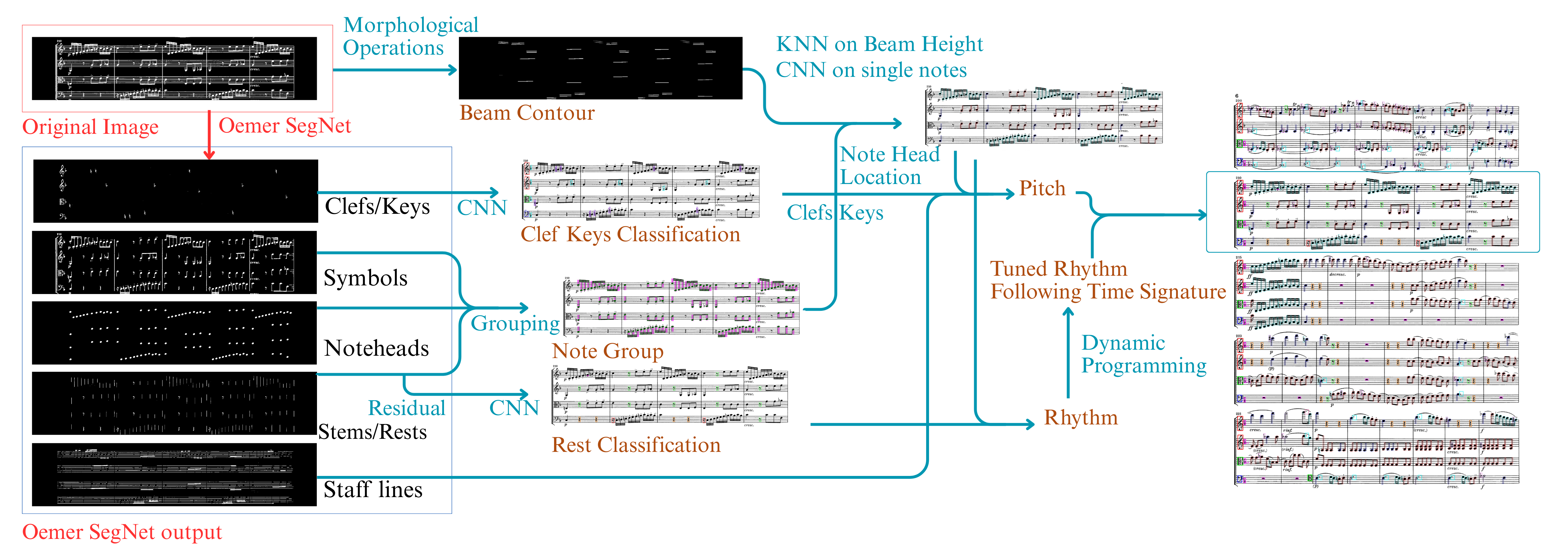}
  \caption{The workflow of the proposed OMR system for string quartets.}
\end{figure*}

\section{Data}

The initial development of our system relied on a set of 20 single-page score images drawn from string and piano sheet music. These images were primarily used for validating implementation logic (e.g., bounding box grouping, and beam parsing) and for iterative refinement of system components. For training the CNN classifiers (e.g., for clefs, accidentals, and rests), candidate bounding boxes were first extracted using contour detection on segmentation maps, and then manually labeled. Since these development images were reused for both tuning and verification, no explicit train/test split was applied at this stage.
To 
avoid overfitting 
on development data, we selected an additional set of 20 unseen images 
for qualitative testing. These pages included varied notation patterns and layout features to verify the robustness of the pipeline. 

We tested the system on the full set of Beethoven’s 16 string quartets from the ABC dataset. These PDF scores were converted into high-resolution page images and processed without any manual intervention, serving as a real-world evaluation of system scalability and symbolic consistency across multiple instrument lines.

\section{Method}\label{sec:page_size}

Our system builds upon the segmentation model, SegNet, by Oemer, which produces five binary maps indicating the locations of 1) stems, 2) clefs and keys, 3) staff lines, 4) noteheads, and 5) other symbols.
From these binary maps, we extract individual musical elements for subsequent integration and rhythm analysis (see Figure 1 for a illustration of the proposed workflow):

\textbf{Clefs and accidentals} are detected via contours from the SegNet model and classified using a CNN. For key signatures, 
we use a pattern-matching algorithm to separate and label individual symbols. 
Due to segmentation limitations for flats and alto clefs, we introduce a YOLO-based detector, which performs more reliably. Future work aims to replace the CNN pipeline entirely with YOLO.

\textbf{Barlines} are identified by scanning vertical lines across the full staff height and applying a pixel-density threshold.

\textbf{Rests} are extracted from the segmentation maps and classified via CNN.

\textbf{Notes} are processed as follows: Noteheads are isolated by filtering the segmentation output; stems are located from notehead corners; pitch is assigned based on vertical position and line intersection. Dots are detected through blob analysis and associated with the nearest notehead. Chords are identified via horizontally aligned notes within stems. Single notes without beams are rhythmically classified using a CNN.

\textbf{Beams} were identified by 
morphological operations 
to the original image to isolate their bounding boxes. Rhythmic values were estimated using two distinct methods based on note grouping. For isolated notes not connected to others via beams, a CNN was used to predict rhythmic duration. For groups of notes connected by beams, rhythmic length was inferred based on the vertical height of the beam structure. We employed KNN clustering on beam height measurements across the image to identify representative groupings, which were then mapped to standard rhythmic values (e.g., eighth, sixteenth, thirty-second) based on common music notation conventions.


Two post-processing stages refine the musical structure:

\textbf{Key signature validation} groups isolated accidentals and matches their layout to known key patterns. Majority vote across all four parts ensures quartet-wide consistency.

\textbf{Rhythmic correction} is performed using a two-layer dynamic programming (DP) algorithm. The first layer finds the best segmentation of notes into durations, minimizing a cost based on (1) notational conventions, (2) deviation from initial predictions, and (3) alignment with expected groupings under the time signature. The second layer assigns durations to notes within each group. If the total duration exceeds the assigned span while all notes have the same class, we infer that tuplets are present. 

\section{results and discussion}\label{sec:typeset_text}


\begin{table}[t]
\centering
\label{tab:overall_accuracy}
\begin{tabular}{lcccccc}
\toprule
Metric & psda & psd & pda & pd & pa & p \\
\midrule
Acc (\%) & 72.1 & 72.2 & 74.6 & 74.7 & 84.2 & 84.3 \\
\bottomrule
\end{tabular}
\caption{OMR performance across all Beethoven quartets (241,301 notes in total), grouped by evaluation matrix.}
\end{table}

We evaluated the output of our optical music recognition system using the ABC dataset as ground truth. Accuracy was measured using a dynamic programming approach that computes the longest common subsequence (LCS) between the predicted and reference note sequences. Each note was represented using combinations of the following four features: \textbf{p} (pitch), \textbf{s} (onset position relative to the bar), \textbf{d} (duration), and \textbf{a} (accidentals).
We defined six evaluation configurations by combining these features: \textit{psda}, \textit{psd}, \textit{pda}, \textit{pd}, \textit{pa}, and \textit{p}. In this paper, we focus on three representative matrices: \textit{psda}, \textit{pda}, and \textit{pa}. The difference in accuracy between matrices that include accidentals and those that do not was found to be less significant (less than 0.1\%), so matrices without accidentals are excluded from visualization for clarity. Results are shown in Table 1.

Across the 16 Beethoven string quartets 
the \textit{psda} configuration achieved an average accuracy of \textbf{72.1\%}, with a maximum of \textbf{88.6\%} (Beethoven 5), a minimum of \textbf{44.3\%} (Beethoven 12), and a median of \textbf{77.8\%}. In comparison, the \textit{pa} configuration yielded a higher average accuracy of \textbf{84.2\%}, with a maximum of \textbf{94.7\%} (Beethoven 5), a minimum of \textbf{73.8\%} (Beethoven 14), and a median of \textbf{84.4\%}.
These results suggest that rhythmic and temporal information (onset and duration) are more difficult to recognize accurately than pitch or accidentals. The substantial improvement in accuracy when excluding these dimensions (from \textit{psda} to \textit{pa}) indicates that rhythm-related errors are the primary contributor to overall recognition challenges.


\bibliography{ISMIRtemplate}

%
%
%
%

\end{document}